\documentclass[letterpaper, 10 pt, conference]{ieeeconf}  % Comment this line out if you need a4paper

\IEEEoverridecommandlockouts                              % This command is only needed if 
\usepackage{graphics} % for pdf, bitmapped graphics files
\usepackage{epsfig} % for postscript graphics files
\usepackage{mathptmx} % assumes new font selection scheme installed
\usepackage{times} % assumes new font selection scheme installed
\usepackage{amsmath} % assumes amsmath package installed
\usepackage{amssymb}  % assumes amsmath package installed
\usepackage{graphicx}
\usepackage{stfloats}    % 允许 \begin{figure*}[!b]

\usepackage{graphicx}
\usepackage{caption} % 提供 \captionof
\usepackage{cuted}
\usepackage{booktabs}

\usepackage{pifont}
\newcommand{\cmark}{\ding{51}}
\newcommand{\xmark}{\ding{55}}

\usepackage[table]{xcolor}  % 提供 \rowcolor，并支持 gray!20 这类颜色混合

\title{\LARGE \bf
GRS-SLAM3R: Real-Time Dense SLAM with Gated Recurrent State
}
\author{Guole Shen\textsuperscript{1,*}, Tianchen Deng\textsuperscript{1,*}, Yanbo Wang\textsuperscript{1}, Yongtao Chen\textsuperscript{1}, Yilin Shen\textsuperscript{1}, Jiuming Liu\textsuperscript{1}, Jingchuan Wang\textsuperscript{1}%
\thanks{Guole Shen, Tianchen Deng, Yanbo Wang, Yongtao Chen, Yilin Shen, Jiuming Liu, Jingchuan Wang are with the Institute of Medical Robotics,  School of Automation and Intelligent Sensing, Shanghai Jiao Tong University, Shanghai 200240, Key Laboratory of System Control and Information Processing, Ministry of Education of China, Shanghai 200240.  The first two authors contribute equally to this paper. (*corresponding author:~jingchuan.wang@sjtu.edu.cn) }%
\thanks{}%
\\
\textsuperscript{1}Shanghai Jiao Tong University
}

\begin{document}

\maketitle
\thispagestyle{empty}
\pagestyle{empty}

\begin{strip}
\vspace*{-2cm} %
  \centering
  \includegraphics[width=\textwidth]{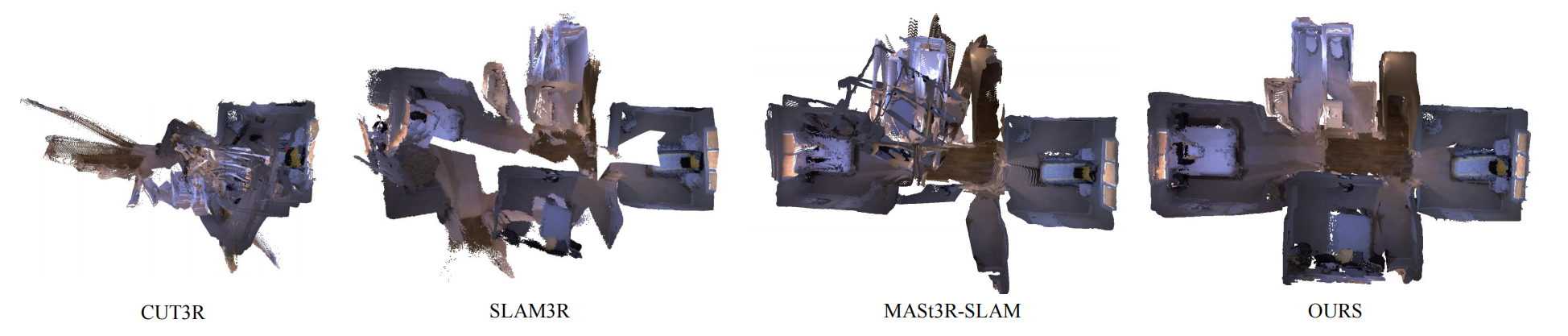}
  \captionof{figure}{Scene Reconstruction Performance. We demonstrate the effectiveness of our method in a large-scale, multi-room apartment scene (\(\sim 100\,\mathrm{m}^2\)). Our approach outperforms existing DUSt3R-based methods in terms of reconstruction quality and completeness.}
  \label{fig:apart0}
\vspace*{-0.3cm} 

\end{strip}

\begin{abstract}

 DUSt3R-based end-to-end scene reconstruction has recently shown promising results in dense visual SLAM. 
 However, most existing methods only use image pairs to estimate pointmaps, overlooking spatial memory and global consistency.
 To this end, we introduce GRS-SLAM3R, an end-to-end SLAM framework for dense scene reconstruction and pose estimation from RGB images without any prior knowledge of the scene or camera parameters.
 Unlike existing DUSt3R-based frameworks, which operate on all image pairs and predict per-pair point maps in local coordinate frames, our method supports sequentialized input and incrementally estimates metric-scale point clouds in the global coordinate.
 In order to improve consistent spatial correlation, we use a latent state for spatial memory and design a transformer-based gated update module to reset and update the spatial memory that continuously aggregates and tracks relevant 3D information across frames. Furthermore, we partition the scene into submaps, apply local alignment within each submap, and register all submaps into a common world frame using relative constraints, producing a globally consistent map. Experiments on various datasets show that our framework achieves superior reconstruction accuracy while maintaining real-time performance.

\end{abstract}

%%%%%%%%%%%%%%%%%%%%%%%%%%%%%%%%%%%%%%%%%%%%%%%%%%%%%%%%%%%%%%%%%%%%%%%%%%%%%%%%
\section{INTRODUCTION}

Dense visual SLAM has been a long-standing challenge in computer vision, aiming to reconstruct the scene and estimate camera poses directly from image inputs.
Over the years, several traditional methods~\cite{orbslam2,deng,xie} have been developed, relying on handcrafted descriptors for image matching and representing scenes with sparse feature point maps. However, the sparsity of these maps makes it difficult for humans to interpret how the system perceives and interacts with the environment. Moreover, such representations fall short of the requirements for tasks like collision avoidance and motion planning. To address these limitations, research has shifted toward dense scene reconstruction, as demonstrated by systems like DTAM~\cite{dtam} and  Kintinuous~\cite{Kintinuous}. Despite their progress, these dense methods often suffer from high memory consumption and slow processing speeds, limiting their applicability in real-time and large-scale scenarios.
Researchers then focus on integrating implicit scene representation~\cite{NeRF} and 3D Gaussian Splatting~\cite{3dgs} with SLAM systems, leading to the emergence of NeRF-based SLAM methods~\cite{niceslam,plgslam,neslam,snislam,deng2025mcnslammultiagentcollaborativeneural} and 3DGS-based SLAM methods~\cite{gsslam,splatam,compact,sgsslam, mgslam}.
However, all these methods rely on accurate camera intrinsics and depth image inputs, which are often difficult to obtain in real-world scenarios.

DUSt3R~\cite{dust3r} shows unprecedented performance and generalization across various real-world scenarios. It operates on image pairs and
uses a global alignment method to align the predicted pointmap into the global map. Some works like CUT3R~\cite{cut3r} and Spann3R~\cite{spann3r} further improve the framework with a continuous update state and an external memory database.
% limit its ability for real-time incremental reconstruction and
% scalability to many images
With the introduction of  DUSt3R~\cite{dust3r}, researchers have begun to combine the DuSt3R framework into dense SLAM. SLAM3r~\cite{slam3r} and MASt3R-SLAM~\cite{mast3rslam} is the pioneer work of DUSt3R-based SLAM. Specifically, SLAM3R proposes a multi-frame registration framework consisting of the Image-to-Points (I2P) network and the Local-to-World (L2W) network, while MASt3R-SLAM leverages the prior from MASt3R and introduces a feature matching based SLAM pipeline. However, these approaches overlook the importance of multi-frame spatial correlation and spatial memory that are crucial for large-scale and long-sequence scene reconstruction.

To address this, we propose an incremental SLAM framework that introduces a latent state to enable metric-scale point cloud estimation, with all reconstructions consistently aligned in the world coordinate system. Importantly, directly updating the latent state with each incoming frame may introduce accumulated drift and noise, particularly under significant viewpoint changes. To mitigate this, we design a gated recurrent mechanism and innovatively introduce two transformer-based gating units: an update gate and a reset gate, which respectively select the relevant information from the current frame, and discard the irrelevant information from the memory.
Furthermore, we use a keyframe-based submap representation with per-submap state reset to reduce drift; inter-submap registration and local refinement aligns submaps while preserving global consistency and local accuracy. \textbf{Overall, our contributions are shown as follows:}
\begin{itemize}
    \item We propose GRS-SLAM3R, a novel end-to-end incremental dense SLAM framework with gated recurrent model and hierarchical submap alignment, achieving accurate scene reconstruction and pose estimation.
    
    \item A novel latent state with gated recurrent model is proposed for consistent spatial correlation. The gated recurrent model includes two transformer-based gates for updating and resetting the memory.
    
    \item We propose a multi-submap scene representation with hierarchical alignment: intra-submap local refinement and inter-submap registration that stitches submaps into a coherent map. Experiments on various datasets demonstrate the superiority of the proposed method in both mapping and tracking.
\end{itemize}

\label{headings}

\begin{figure*}[t]
  \centering
    \includegraphics[width=\linewidth]{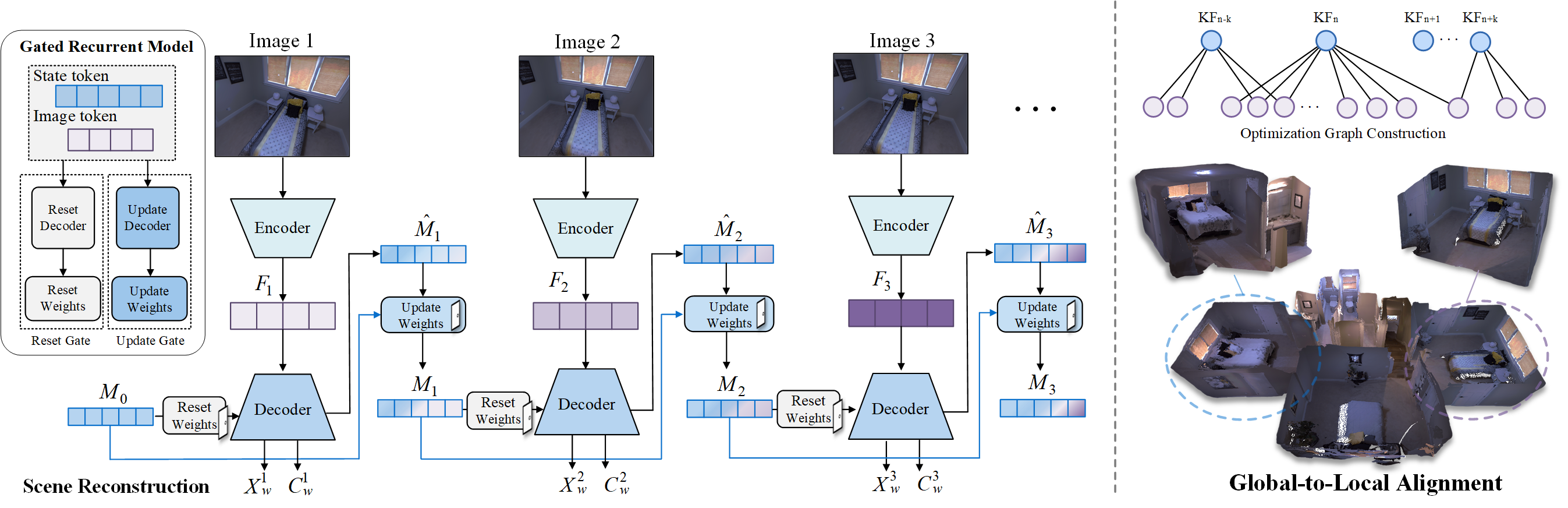}

    \caption{\textbf{System overview. }The input to our system is a stream of RGB images, and the output of the GRS-SLAM3R system consists of camera poses and dense point clouds. On the left side of the figure, we illustrate our scene representation based on a gated recurrent mechanism. Each input frame is encoded into image tokens via an encoder, which interact with a latent state $M_t$ that is updated through reset and update gates. The updated state $\hat{M_t}$ is then decoded to produce the corresponding pose and point cloud in the world coordinate system. On the right side, we present the multi-submap scene representation.}
 
    \label{fig:framework}
    \vspace*{-0.5cm} 
\end{figure*}

\section{Related work}

\subsection{Dense Monocular SLAM}
Dense monocular SLAM aims to reconstruct geometry using only RGB images, avoiding the need for depth sensors. Early methods optimize photometric consistency between frames to estimate depth and pose~\cite{dtam,lsd-slam}, but were limited by drift, noise, and incomplete reconstructions. To enhance robustness and completeness, later approaches incorporate learned priors for depth and visual features~\cite{cnn-slam,codeslam}, combining data-driven inference with geometric optimization. This hybrid strategy improves reconstruction quality but introduces runtime overhead and generalization issues. More recent systems tightly couple learning and optimization~\cite{droidslam, tandem}, achieving better accuracy, though dense monocular SLAM still struggles with real-time performance and global consistency in unconstrained environments.
\vspace{-0.1cm}
\subsection{3D Reconstruction}
Dense 3D reconstruction remains a fundamental yet challenging task in computer vision. Traditional methods follow a sequential pipeline: Structure-from-Motion (SfM)~\cite{pixelperfectsfm, Buildingromeinaday, sfmrevisited, optimizingsfm, globalsfmrevisit, Phototourism} recovers sparse geometry and camera poses through keypoint detection, matching, triangulation, and bundle adjustment. Multi-View Stereo (MVS)~\cite{mvsnet, mvstutorial, MVSeval, accuratemvs, pixelwisemvs} estimates dense geometry from known poses. While recent works improve individual components with learned features~\cite{superpoint}, neural bundle adjustment~\cite{ba-net}, and explore depth prediction~\cite{unidepth,zoedepth,megadepth,diggingmonodepth} as an alternative means to enhance geometry estimation, these pipelines remain sensitive to early-stage errors and typically require camera calibration and offline optimization. Neural rendering methods like NeRF~\cite{NeRF} and 3D Gaussian Splatting~\cite{3dgs} offer high fidelity but still rely on known poses and per-scene optimization.

\subsection{DUSt3R-Based Online Continuous 3D Reconstruction}
DUSt3R~\cite{dust3r} introduces a groundbreaking paradigm shift by directly regressing a pointmap, a standard representation in visual localization, from a pair of images without relying on any prior knowledge of the scene. This approach challenges conventional assumptions and opens up new possibilities for dense reconstruction from minimal input. Several recent works extend DUSt3R’s pointmap-based reconstruction to online and continuous settings. Spann3R~\cite{spann3r} uses an external spatial memory to perform incremental scene reconstruction in a unified coordinate system. CUT3R~\cite{cut3r} incorporates recurrent states for sequential integration. However, these methods often suffer from drift and geometric inconsistency due to the lack of global correction. SLAM3R~\cite{slam3r} and MASt3R-SLAM~\cite{mast3rslam} are the most relevant SLAM systems to our work. SLAM3R performs incremental mapping through the Local-to-World (L2W) network, while MASt3R-SLAM builds on the MASt3R~\cite{mast3r} two-view prior and integrates efficient point map matching, tracking, fusion, and global optimization. However, both methods overlook the spatial memory and multi-frame correlation, resulting in a lack of consistency during the mapping process.

\section{Method}

In this paper, we propose GRS-SLAM3R, a novel monocular RGB SLAM system for high-quality online dense 3D reconstruction.

Our system processes a sequence of monocular RGB images $\left\{I_t \in \mathbb{R}^{H \times W \times 3}\right\}_{t=1}^N$ without relying on external pose or depth supervision. The output of our system is the dense 3D pointcloud $X_t \in \mathbb{R}^{M \times 3}$, where M is the number of 3D points, the corresponding confidence $C_t$, and the pose $P_t$.
%%%%太具体了，适合在method讲，不适合在overview
We show the overview of our system in Fig.~\ref{fig:framework}. We design a persistent latent state representation with gated update model for spatial correlation and long-range memory (Sec.~\ref{sec:mapping}). Furthermore, We use a submap-reset representation that updates the latent state locally, preventing state degradation and reducing long-range drift. To integrate multiple submaps, we adopt inter-submap registration and intra-submap refinement.(Sec.~\ref{sec:align}).

% Specifically, the image simultaneously updates the
% state with new information and retrieves information stored
% in the state. Following the state-image interaction, explicit
% 3D pointmaps and camera poses are extracted for each view
%spatial memory

% We maintain a persistent latent state representation that continuously evolves with the incoming images. We design spatially-varying update and reset gates to control the integration of new information and the removal of outdated memory. Explicit 3D point clouds and camera poses are predicted after each state-image interaction.

%%%%太具体了，适合在method讲，不适合在overview
% To further maintain global structural consistency and achieve fine-grained reconstruction accuracy, we introduce a hierarchical optimization strategy. Keyframes are first globally aligned into a coherent world coordinate system, correcting large-scale drift. Local alignment focuses on enhancing the geometric consistency of keyframe pairs within each submap, leveraging their fixed global poses for accurate reconstruction.

\subsection{Scene Reconstruction}
\label{sec:mapping}
Existing approaches like DUSt3R~\cite{dust3r} operate on pairs of images, which limits their scalability and suitability for real-time, incremental reconstruction. To address this, we propose an incremental scene representation framework with a latent state. Specifically, we design a gated update mechanism that employs two transformer-based gates to effectively update the latent state of the current frame while discarding irrelevant information, thereby preventing the introduction of noise into the memory. In contrast to pairwise predictions in DUSt3R~\cite{dust3r}, our method can directly output point maps in the world coordinate, which better fits the requirements of online SLAM systems.

% Unlike DUSt3R, which
% predicts per image-pair pointmaps each expressed in its
% local coordinate frame,  can predict per-image
% pointmaps expressed in a global coordinate system, thus
% eliminating the need for optimization-based global alignment. The key idea of is to manage an external
% spatial memory that learns to keep track of all previous relevant 3D information. then queries this spatial
% memory to predict the 3D structure of the next frame in a
% global coordinate system.
%%应该要先讲motivation，设计思路，既有cut3r也有gated 首先需要满足SLAM的incremental的性质

% However, operating on a pair of images and
% the need for per-scene optimization-based global alignment
% limit its ability for real-time incremental reconstruction and
% scalability to many images.
%现有的dust3r方法，包括slam3r 都是pair的输入，我们首先改成了incremental的场景表示方法。同时设计了一个Gate updated的state来与the incoming images交互，从而实现predict the 3D structure of the next frame in a global coordinate system. 相比于dust按照pair的表示，直接输出世界坐标系的地图更符合SLAM系统的需求。
%\vspace{-0.3cm}
\subsubsection{Gated Recurrent Model}
At each time step $t$, the system receives an input image $I_t$, which is first encoded into token representations $F_t$ through a vision transformer encoder \cite{vit}:
\begin{equation}
F_t = \text{Encoder}(I_t)
\end{equation}
%在这里面，讲到持续更新的state之后，引入cut3r，和他对比，提出改进的motivation。
Inspired by CUT3R~\cite{cut3r}, we incorporate a latent memory state $M_t$ that progressively accumulates scene information across sequential observations. We represent the latent state as a set of tokens.
% 直接进行state的interact with image,同时进行state的更新can lead to drift, noise amplification, and contamination of long-term memory.

However, directly integrating new image features $F_t$ into the state without regulation can result in drift, the introduction of noise, and contamination of long-term memory, which becomes particularly problematic in long sequences. To address these challenges, we design a gated update mechanism inspired by the traditional GRU structure~\cite{GRU}, which selectively integrates new information while preserving existing memory. The image tokens $F_t$ will interact with the memory in two directions. We update the state with information from the current image feature, then we retrieve contextual cues from the state to incorporate knowledge accumulated from past frames. This mechanism ensures stable state updates and accurate information propagation, enabling effective spatial correlation across multiple frames and consistent scene reconstruction over long sequences. 
%这个机制可以保证稳定的state更新，更加准确的state信息传递和空间多帧的correlation，保证consistent scene reconstruction over the long-range sequence.
%GRU 有两个有两个门，即一个重置门（reset gate）和一个更新门（update gate）。从直观上来说，重置门决定了如何将新的输入信息与前面的记忆相结合，更新门定义了前面记忆保存到当前时间步的量。为了解决标准 RNN 的梯度消失问题，GRU 使用了更新门（update gate）与重置门（reset gate）。基本上，这两个门控向量决定了哪些信息最终能作为门控循环单元的输出。这两个门控机制的特殊之处在于，它们能够保存长期序列中的信息，且不会随时间而清除或因为与预测不相关而移除。更新门帮助模型决定到底要将多少过去的信息传递到未来，或到底前一时间步和当前时间步的信息有多少是需要继续传递的。这一点非常强大，因为模型能决定从过去复制所有的信息以减少梯度消失的风险。我们随后会讨论更新门的使用方法，现在只需要记住 z_t 的计算公式就行。
%重置门主要决定了到底有多少过去的信息需要遗忘，我们可以使用以下表达式计算：现在我们应该比较了解到底 GRU 是如何通过更新门与重置门存储并过滤信息。门控循环单元不会随时间而清除以前的信息，它会保留相关的信息并传递到下一个单元，因此它利用全部信息而避免了梯度消失问题
% The gated state update mechanism contains two stages: a) memory suppression via a reset gate. b)controlled integration via an update gate.

\begin{figure}
  \centering
  \includegraphics[width=0.8\linewidth]{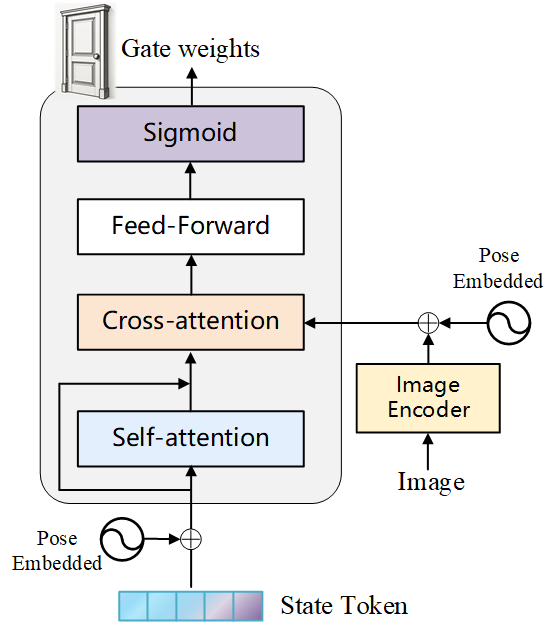}

  \caption{\textbf{Gate structure}. We illustrate the detailed structures of the reset and update gates we designed. Both gates take the historical latent state and the current frame as inputs, and compute the gating weights through a combination of self-attention and cross-attention mechanisms.}
  \label{fig:gate}

\end{figure}

To this end, we design two transformer-based gates in the decoder: a reset gate $G_r(\cdot)$ and an update gate $G_u(\cdot)$. The reset gate determines how the current input information is combined with the previous memory, while the update gate controls how much of the previous memory is preserved and carried over to the current time step.
%%%介绍这个reset门，说清楚输入输出，网络结构，是借鉴的dust3r的deocder的transformer结构可以结合，image和state的corss attention
The detailed structure of the gates is shown in Fig.~\ref{fig:gate}. 
The input of the reset gate $G_r(\cdot)$ is image tokens of the current frame $F_t$ and the memory state of the last time step $M_{t-1}$. The output of the reset gate $U_t$ determines which parts of the previous memory  $M_{t-1}$  should be suppressed before incorporating new observations, thereby regulating the influence of outdated or irrelevant information. The gates can be formulated as: 
\begin{equation}
R_t = G_r(M_{t-1}, F_t), \quad U_t = G_u(M_{t-1}, F_t)
\end{equation}
The input of update gate $G_u(\cdot)$ is the same as the reset gate. The update gate controls how much information $M_{t-1}$ should be carried forward, determining the extent to which the memory from the previous timestep is preserved. 
The output of the reset gate $G_r$ is then applied element-wise to the memory:
\begin{equation}
M^{\text{reset}}_t = R_t \odot M_{t-1}
\end{equation}

Then, we obtain the reset memory $M^{\text{reset}}_t$ where irrelevant or outdated information is discarded. To integrate the updated memory with current observations, we design a transformer-based decoder, in which the memory tokens interact with the current frame features $F_t$ and the pose token $z_t$:
\begin{equation}
[\hat{M}_t, z'_t\oplus F'_t ] = \text{Decoder}\left([M^{\text{reset}}_t, z_t \oplus F_t]\right)
\end{equation}
where $\hat{M}_t$ denotes updated memory state, $F'_t$ are the enriched frame features, and $z'_t$ is a dedicated pose token capturing global scene-level information. The decoder applies cross-attention between memory tokens and image tokens at each block, enabling bidirectional information flow across memory, image, and pose tokens.

The final memory state is then updated via gated update:
\begin{equation}
M_t = U_t \odot \hat{M}_t + (1 - U_t) \odot M_{t-1}
\end{equation}

This two-stage gating process enables our system to selectively suppress unreliable memory content while adaptively integrating informative new observations, enhancing long sequence consistency and robustness.

After the gated memory update, explicit metric-scale 3D pointclouds can
be extracted from $z'_t$ and $F'_t$ for each frame. Specifically, we can generate dense 3D point clouds in both the local camera frame and the global world frame, along with 6-DoF camera poses:
\begin{equation}
\hat{X}^{\text{self}}_t, \hat{C}^{\text{self}}_t = \text{Head}_{\text{self}}(F'_t)
\end{equation}
\begin{equation}
\hat{X}^{\text{world}}_t, \hat{C}^{\text{world}}_t = \text{Head}_{\text{world}}(F'_t, z'_t)
\end{equation}
\begin{equation}
\hat{P}_t = \text{Head}_{\text{pose}}(z'_t)
\end{equation}
where $\text{Head}_{\text{self}}(\cdot)$ and $\text{Head}_{\text{world}}(\cdot)$ are implemented as DPT~\cite{dpt},
and $\text{Head}_{\text{pose}}(\cdot)$ is implemented as an MLP network, respectively. All pointmaps and poses are in metric scale.

%%%需要在这个其实总结一下这个段落，然后大概聊一下motivation
% keyframe选择、子地图重置state？
\subsection{Hierarchical Submap Alignment}
\label{sec:align}
While the gated memory mechanism facilitates the selection of historical memories, it suffers from accumulated drift in long-sequence. To address this issue, we propose a multi-submap scene representation method, where submaps are segmented based on changes in keyframe viewpoints. We adopt a hierarchical alignment scheme: local alignment within each submap and inter-submap registration for global consistency. This two-level process limits error accumulation over long sequences and preserves fine-grained geometric accuracy, improving cross-submap coherence.

\paragraph{Frontend Submap Construction and Keyframe Selection}
As each new frame $I_t$ arrives, we evaluate covisibility with the last keyframe $I_k$ and the current submap’s anchor $I_{a_s}$ (the first frame of submap $S_s$). If $\mathrm{Cov}(I_t,I_k)<\tau_{\text{kf}}$, we promote $I_t$ to a keyframe within $S_s$. If $\mathrm{Cov}(I_t,I_{a_s})<\tau_{\text{anchor}}$, we start a new submap $S_{s+1}$, reset the latent state, and set $I_t$ as its anchor. Each submap maintains its own latent memory and local coordinate frame. This submap–reset policy bounds error accumulation within a submap and prevents long-horizon drift from propagating through the recurrent state.

\paragraph{Local Submap Alignment}

We perform local alignment within each submap. 
We construct a local connectivity graph $\mathcal{G}_l = (\mathcal{V}_l, \mathcal{E}_l)$ for each submap. The vertex set $\mathcal{V}_l$ includes all frames in the submap, and the edge set $\mathcal{E}_l$ is built by treating each keyframe $k \in \mathcal{K}_l$ as a center and connecting it with its temporally adjacent keyframes and ordinary frames within a fixed window. Specifically, for a keyframe $n$, we define a local neighbor $\{n-k, \dots, n+k\}$ and construct pairwise edges with all frames in this group. Each edge is associated with predicted dense pointmaps and confidence maps, and contributes to the local alignment loss.  We use local alignment to register all frames within each submap to the world coordinate system, jointly optimizing the pose and point cloud of each frame.
\begin{equation}
\mathcal{L}_{\text{local}}^{S}
= \sum_{e \in \mathcal{E}_l^{S}} \sum_{v \in e} \sum_{i=1}^{HW}
C_i^{v,e}\,\big\|\, \xi_i^{v,S} - \sigma_e^{S}\, P_e^{S}\, X_i^{v,e} \big\|
\end{equation}
where $\xi^{v,S}$ denotes the optimized local pointmap for keyframe $v$ in submap $S$, $P_e^{S}$ is the per-edge rigid transform that maps $X^{v,e}$ to the submap coordinate, and $\sigma_e^{S}\!>\!0$ is an optional per-edge scale. Minimizing $\mathcal{L}_{\text{local}}^{S}$ jointly refines the poses and pointmaps of frames within $S$, producing a self-consistent local reconstruction.

\paragraph{Inter-Submap Alignment}

We construct a pose graph with submap poses $T_s\in\mathrm{SE}(3)$ as nodes and edges from adjacent-submap alignments and loop closures,
 Because all submaps share a common metric scale, the inter-submap constraint is a pure rigid transform $\Delta T_{s,s+1}\in\mathrm{SE}(3)$. When $S_s$ is finalized, we query all previous keyframes and retain the top-ranked loop hat exceeds the score threshold, yielding additional relative constraints $\Delta T_{i,j}$.  and then solve the resulting pose-graph optimization.
\begin{equation}
\resizebox{.8\columnwidth}{!}{$
\min_{\{T_s\}}\ \sum_{(u,v)\in\mathcal{E}}
 \bigl\|\operatorname{Log}(\Delta T_{u,v}^{-1} T_u^{-1} T_v)\bigr\|_{\Sigma}^{2}
 + \bigl\|\operatorname{Log}(T_{s_0}^{-1}\bar T_{s_0})\bigr\|_{\Sigma_0}^{2}
$}
\end{equation}
where the small-noise prior $(s_0,\bar T_{s_0},\Sigma_0)$ fixes the global gauge. Constraints are added incrementally, adding a loop appends a factor and triggers Levenberg--Marquardt re-optimization.

\subsection{Training loss}
Given a sequence of $N$ images, we follow CUT3R~\cite{cut3r} and supervise with a confidence-weighted 3D regression loss and a pose loss. 

\paragraph{Regression loss}
We supervise dense pointmap prediction with a scale-aware, confidence-weighted loss. Let $x_i$ be the ground-truth 3D point and $\hat{x}_i$ the prediction at pixel $i$ (confidence $c_i\in[0,1]$). Denoting normalization factors by $s$ (GT) and $\hat{s}$ (prediction), we define
\begin{equation}
\mathcal{L}_{\text{regr}}=\sum_{i=1}^{M}\Big(c_i\big\|\tfrac{\hat{x}_i}{\hat{s}}-\tfrac{x_i}{s}\big\|_2-\beta\log c_i\Big).
\end{equation}
When the ground-truth pointmaps are already in metric scale, we set $\hat{s}=s$ to remove scale ambiguity.

\paragraph{Pose loss}
Let the predicted pose be $\hat{P}_t=(\hat{q}_t,\hat{\tau}_t)$ with quaternion $\hat{q}_t$ and translation $\hat{\tau}_t$ at time $t$; $(q_t,\tau_t)$ are the ground-truths. We minimize rotational and translational discrepancies (the latter normalized consistently with the regression loss):
\begin{equation}
\mathcal{L}_{\text{pose}}=\sum_{t=1}^{N}\left(\|\hat{q}_t-q_t\|_2+\big\|\tfrac{\hat{\tau}_t}{\hat{s}}-\tfrac{\tau_t}{s}\big\|_2\right).
\end{equation}

\paragraph{Curriculum training}
We resize the longer image side to 512 and adopt a three-stage curriculum for stable convergence. We adopt a three-stage curriculum: first, we train on 4-frame sequences, freezing the encoder and part of the decoder and updating only the two gating modules; second, we keep the encoder frozen, unfreeze the decoder, and train the decoder together with the gating modules; finally, we extend sequences to 64 frames and fine-tune the decoder, gating modules, and prediction heads.The latter stages strengthen inter-frame reasoning and improve long-range spatial modeling. We train our network on a various
set of 10 datasets, covering synthetic and real-world
data, scene-level and object-centric scenes, as well as both indoor and outdoor
scenes. Examples of our datasets include CO3Dv2~\cite{co3dv2}, ARKitScenes~\cite{arkitscenes}, ScanNet~\cite{scannet}, WildRGBD~\cite{wildrgbd}, BlendedMVS~\cite{blendedmvs}, Matterport3D~\cite{matterport3d}.

\section{Experiments}

\label{sec:exp}

\subsection{Experimental Setup}
 We evaluate our method in terms of surface reconstruction quality, camera pose estimation accuracy, and real-time performance.
\paragraph{Test Datasets}
We evaluate the effectiveness of our method on small-scale real-world scenes using NRGBD~\cite{neural_rgbd} and a subset of 18 sequences from 7-Scenes~\cite{7scenes}.
To further demonstrate performance in long sequences and large-scale scenarios, we evaluate on the Apartment dataset~\cite{replica} (multi-room, $\sim\!100\,\mathrm{m}^2$) and on the NES dataset~\cite{deng2025mne}~\cite{deng2025mcnslammultiagentcollaborativeneural}. 
NES spans $>1000\,\mathrm{m}^2$ with a total trajectory length of $1{,}482.75\,\mathrm{m}$.

\paragraph{Evaluation Metrics} We use absolute trajectory error (ATE-RMSE) to evaluate camera tracking. We evaluate reconstruction quality using accuracy and completeness. Following NICER-SLAM~\cite{nicerslam}, Spann3R~\cite{spann3r}, and SLAM3R~\cite{slam3r}, we generate ground-truth point clouds by projecting depth maps into 3D using known camera intrinsics and poses for each test sequence. To account for potential scale discrepancies across methods, we adopt the alignment strategy of SLAM3R~\cite{slam3r}: a global similarity transform estimated via the Umeyama algorithm, followed by ICP refinement to minimize residual geometric error. 

\paragraph{Implementation Detail}
 All the training experiments are conducted on eight NVIDIA A100 GPUs with 80 GB of memory each. The inference and SLAM experiments are conduct on a RTX 4090 GPU.

\subsection{Camera Pose Estimation}

\begin{figure*}[!t]
    \centering
    \includegraphics[width=\linewidth]{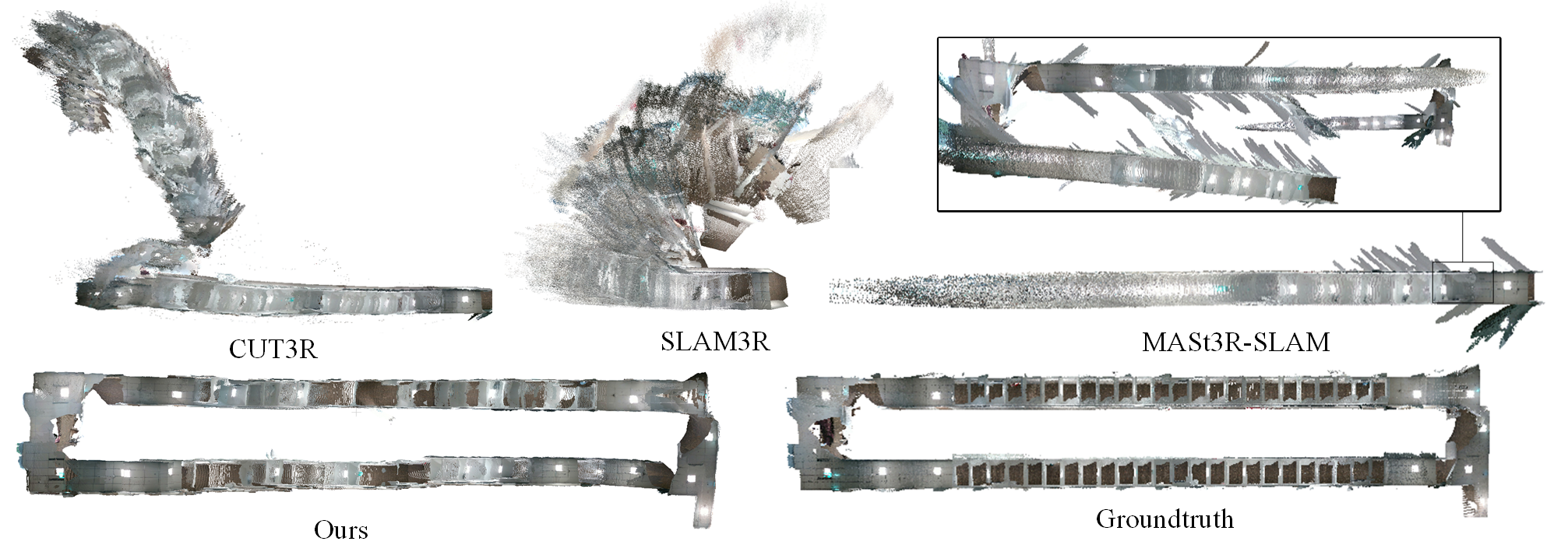} 
    \caption{\textbf{Qualitative scene reconstruction results.} We demonstrate the mapping performance of our method on the NES dataset. For MASt3R-SLAM~\cite{mast3rslam}, due to the scale reduction in the latter part of the sequence, we provide an enlarged view of this region alongside the full-scene overview. The boxed region highlights the scale-reduced area.}
    \label{fig:indoor}
\end{figure*}

Our method demonstrates robust and competitive pose estimation performance across both small- and large-scale scenes, particularly when compared with existing DUST3R-based online approaches.
On smaller indoor datasets (e.g., 7-Scenes), the gated recurrent update stabilizes per-frame poses and consistently lowers ATE (Table~\ref{tab:7scenes_ATE}). In challenging large-scale, multi-room scenes such as Apartment, shown in Table~\ref{tab:apart0}, existing methods suffer from drift or trajectory collapse under long-range motion or abrupt camera rotation. Figure~\ref{fig:apart0} illustrates that CUT3R~\cite{cut3r} fails to reconstruct the scene, SLAM3R~\cite{slam3r} and MASt3R-SLAM~\cite{mast3rslam} suffer from significant pose drift in certain rooms, whereas our method maintains consistent and stable tracking throughout the scene. Our multi-submap scene representation method bounds the drift and reduces the risk of accumulated drift over long sequences.  
As shown in Fig.~\ref{fig:indoor}, on complex indoor sequences, CUT3R~\cite{cut3r} drifts after sharp turns, SLAM3R~\cite{slam3r} breaks in long corridors, and MASt3R-SLAM~\cite{mast3rslam} exhibits significant scale variation, with the reconstructed scene size abruptly decreasing after a certain sequence length. In comparison, our method is able to reconstruct corridor structures with stable scale and spatial consistency.

\begin{table}[t]
\centering
\resizebox{\columnwidth}{!}{%
\begin{tabular}{lrrrrrrrr}
\toprule
\textbf{Method} & \textbf{Chess} & \textbf{Fire} & \textbf{Heads} & \textbf{Office} & \textbf{Pump.} & \textbf{RedKit.} & \textbf{Stairs} &\textbf{Avg.}  \\
\midrule
CUT3R~\cite{cut3r}        & 5.90 & 5.34 & 6.37 & 13.85 & 14.73 & 9.44 & \textbf{6.67} & 8.90 \\
MASt3R-SLAM~\cite{mast3rslam}  & 7.24 & 5.78 & \textbf{3.68} & \textbf{13.31} & \textbf{12.87} & 10.07 & 6.68 & 8.52 \\
\textbf{Ours}         & \textbf{5.30} & \textbf{5.31} & 4.09 & 13.43 & 14.41 & \textbf{8.95} & 6.81 & \textbf{8.27}  \\
\bottomrule
\end{tabular}%
}
\caption{Qualitative pose estimation  runtime on 7Scenes dataset~\cite{7scenes}. We report per-scene ATE RMSE in centimeters.}
\label{tab:7scenes_ATE}

\end{table}

\begin{table}[t]
  \centering
  \begin{tabular}{lcccc}
    \toprule
    \textbf{Method} & \multicolumn{2}{c}{\textbf{Acc.}[cm]} & \multicolumn{2}{c}{\textbf{Comp.}[cm]} \\
                    & Mean & Median & Mean & Median \\
    \midrule
    DUSt3R-GA~\cite{dust3r}        & 0.144 & \textbf{0.019} & 0.154 & \textbf{0.018} \\
    MASt3R-GA~\cite{mast3r}        & \textbf{0.085} & 0.033 & \textbf{0.063} & 0.028 \\
    \midrule
    Spann3R~\cite{spann3r}         & 0.416 & 0.323 & 0.417 & 0.285 \\
    CUT3R~\cite{cut3r}             & 0.099 & 0.031 & 0.076 & 0.026 \\
    \textbf{Ours}                  & \underline{0.089} & \underline{0.030} & \underline{0.072} & \underline{0.025} \\
    \bottomrule
  \end{tabular}
  \caption{Reconstruction results on the NRGBD dataset~\cite{neural_rgbd}.}
  \label{tab:nrgbd_results}
  \vspace*{-0.5cm} 
\end{table}

\begin{table*}[t]
\centering
\resizebox{\textwidth}{!}{%
\begin{tabular}{lcccccccc>{\columncolor{gray!12}}c}
\toprule
\textbf{Method} & \textbf{Chess} & \textbf{Fire} & \textbf{Heads} & \textbf{Office} & \textbf{Pumpkin} & \textbf{RedKitchen} & \textbf{Stairs} & \textbf{Average} & \cellcolor{gray!12}\textbf{FPS} \\
& Acc. / Comp. & Acc. / Comp. & Acc. / Comp. & Acc. / Comp. & Acc. / Comp. & Acc. / Comp. & Acc. / Comp. & Acc. / Comp. & \\
\midrule
DUSt3R~\cite{dust3r}       & 2.26 / 2.13 & 1.04 / 1.50 & 1.66 / \underline{0.98} & 4.62 / 4.74 & \textbf{1.73} / 2.43 & \underline{1.95} / 2.36 & 3.37 / 10.75 & 2.19 / 3.24 & $<1$ \\
MASt3R~\cite{mast3r}       & 2.08 / 2.12 & 1.54 / 1.43 & \textbf{1.06} / 1.04 & \underline{3.23} / 3.19 & 5.68 / 3.07 & 3.50 / 3.37 & \underline{2.36} / 13.16 & 3.04 / 3.90 & $<1$ \\
Spann3R~\cite{spann3r}     & 2.23 / 1.68 & \underline{0.88} / \underline{0.92} & 2.67 / \underline{0.98} & 5.86 / 3.54 & 2.25 / \textbf{1.85} & 2.68 / \textbf{1.80} & 5.65 / \underline{5.15} & 3.42 / 2.41 & $>50$\\
CUT3R~\cite{cut3r}         & 2.46 / 1.99 & 1.52 / 1.43 & 2.10 / 1.13 & 3.81 / 3.05 & 2.98 / 2.48 & 2.49 / 2.24 & 3.35 / 10.53 & 2.67 / 3.27 & $\sim 20$ \\
SLAM3R~\cite{slam3r}       & \underline{1.63} / \textbf{1.31} & \textbf{0.84} / \textbf{0.83} & 2.95 / 1.22 & \textbf{2.32} / \textbf{2.26} & \underline{1.81} / \underline{2.05} & \textbf{1.84} / \underline{1.94} & 4.19 / 6.91 & \underline{2.13} / \underline{2.34} & $\sim 25$ \\
MASt3R-SLAM~\cite{mast3rslam} & 2.41 / 1.70 & 1.57 / 1.33 & 1.71 / 1.16 & 3.47 / \underline{2.98} & 2.86 / 2.37 & 2.83 / 2.16 & 3.32 / 9.53 & 2.60 / 3.03 & $\sim 15$ \\
\textbf{Ours}              & \textbf{1.49} / \underline{1.32} & 1.26 / 1.32 & \underline{1.22} / \textbf{0.83} & 4.17 / 3.41 & 2.27 / 2.25 & 2.19 / 2.19 & \textbf{2.22} / \textbf{4.55} & \textbf{2.12 / 2.27} & $\sim 15$ \\
\bottomrule
\end{tabular}
}
\caption{\textbf{Reconstruction results and runtime (FPS) on 7Scenes~\cite{7scenes}}. We report Accuracy and Completion in centimeters. }
\label{tab:7scenes_results}
\end{table*}

\begin{table}[t]
  \centering
  \begin{tabular}{lccc}
    \toprule
    Method & \textbf{Acc.}[cm] $\downarrow$ & \textbf{Comp.}[cm] $\downarrow$ & \textbf{ATE}[m] $\downarrow$ \\
    \midrule
    CUT3R~\cite{cut3r}            & 48.49 & 39.85 & 2.39 \\
    SLAM3R~\cite{slam3r}          & 21.67 & 27.66 & \multicolumn{1}{c}{--} \\
    MASt3R-SLAM~\cite{mast3rslam} & \underline{8.72} & \underline{5.80} & \underline{0.72} \\
    \textbf{Ours}                 & \textbf{6.79} & \textbf{5.62} & \textbf{0.16} \\
    \bottomrule
  \end{tabular}
  \caption{Quantitative results on the Apartment dataset~\cite{replica}. 
  ATE is not reported for SLAM3R~\cite{slam3r} as it does not produce explicit camera pose estimates during inference.}
  \label{tab:apart0}
  \vspace{-0.2cm}
\end{table}

\begin{figure*}[t]
    \centering
    \includegraphics[width=\linewidth]{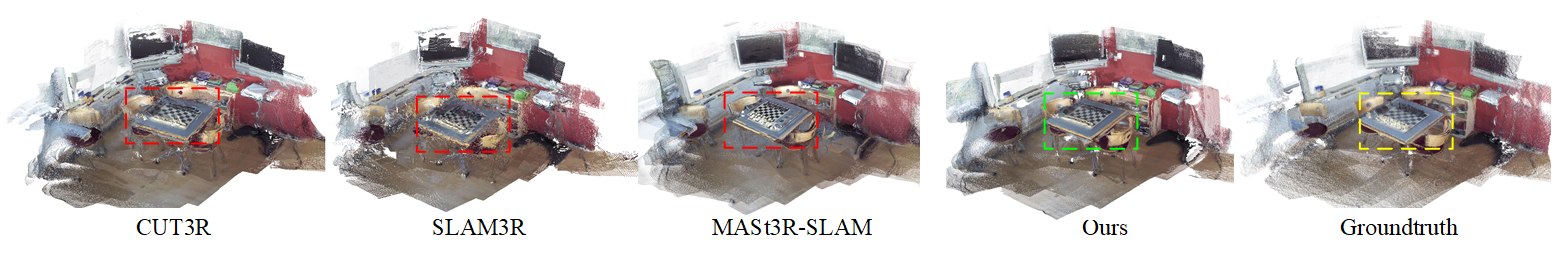} 
    \caption{\textbf{Qualitative scene reconstruction results.} We demonstrate the mapping performance of our method on chess seq-05 in 7scenes~\cite{7scenes}. The region outlined on the image is marked in red to signify lower predictive accuracy, in green to signify higher accuracy, and in yellow to represent the ground truth results.}
    \label{fig:7scenereplica}
\end{figure*}

\subsection{Surface Reconstruction}

Our method achieves high-quality scene reconstruction across different datasets. We follow the~\cite{cut3r} setting to evaluate performance on sparsely sampled images from the NRGBD dataset, which contains minimal or no view overlap, as shown in Table~\ref{tab:nrgbd_results}. The results indicate that the recurrent mechanism preserves key geometry across frames and enables spatial reasoning by selectively integrating structural cues, even with minimal overlap. According to the results in Table~\ref{tab:7scenes_results}, our approach maintains stable and high-quality reconstructions in small indoor scenes such as 7Scenes. Spann3R~\cite{spann3r} and CUT3R\cite{cut3r} both incorporate spatial memory mechanisms, similar to our approach. However, they do not effectively address the challenges of memory update and forgetting, which are crucial for spatial memory consistency and robustness. Our recurrent model helps maintain geometric consistency across frames, while local refinement improves surface alignment and enhances fine-grained details. In large-scale scenes, as shown in Table~\ref{tab:apart0}, Fig.~\ref{fig:apart0} and Fig.~\ref{fig:indoor}, our method achieves accurate and consistent reconstruction, which indicates the effectiveness of our framework. Compared with MASt3R-SLAM~\cite{mast3rslam} and SLAM3R~\cite{slam3r}, our framework builds consistent spatial multi-view correlations through the gated recurrent model, and further enhances reconstruction consistency via hierarchical alignment.

\subsection{Time Analysis}
In Table~\ref{tab:7scenes_results}, we present the runtime analysis of our framework and other methods. Our method runs at approximately 15 FPS on a single RTX 4090 GPU, achieving a favorable balance between speed and accuracy compared to existing online SLAM baselines.

% \begin{table}[t]
%   \centering
%   \small
%   \begin{tabular}{lcc}
%     \toprule
%     \textbf{Methods} & \textbf{Acc./ Comp.}[cm] & \textbf{ATE}[cm] \\
%     \midrule
%     w/o Gated Update     & 5.87 / 5.30 & 6.42 \\
%     w/o Submap           & 3.95 / 2.93 & 6.38 \\
%     w/o Local Align        & 5.57 / 3.88 & 8.45 \\
%     \textbf{Full Model (Ours)} & \textbf{3.34} / \textbf{2.43} & \textbf{5.58} \\
%     \bottomrule
%   \end{tabular}
%   \caption{Ablation study on core components. We conduct experiments on the Replica dataset~\cite{replica} to verify the effectiveness of our method. Our full model achieves better completion reconstructions and more accurate pose estimation results.}
%   \label{tab:ablation}
% \end{table}

\begin{figure}[t]
  \centering
  \includegraphics[width=\linewidth]{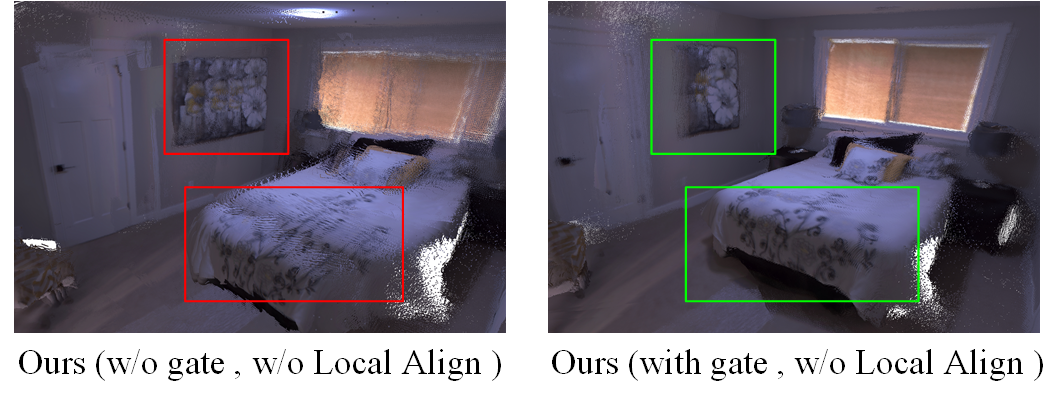}
  \caption{Ablation on Gated Recurrent model. We conduct an ablation study on the Apartment dataset~\cite{replica} to evaluate the effectiveness of the gated recurrent state. The gate update module significantly enhances the accuracy and consistency of scene reconstruction.}
  \label{fig:ablation}
\end{figure}

\begin{table}[t]
\centering
\begin{tabular}{cccccc}
\toprule
\multicolumn{3}{c}{Modules} & \multicolumn{2}{c}{Reconstruction (cm)} & {ATE (m)} \\
\cmidrule(lr){1-3}\cmidrule(l){4-5}
Gate & Local Align & Submap & {Acc.} & {Comp.} & {} \\
\midrule
\xmark & \xmark & \xmark & 48.49  & 39.85 & 2.39 \\
\xmark & \cmark & \cmark & 8.87  & 6.30  & 0.33 \\
\cmark & \cmark & \xmark & 28.95  & 24.93  & 1.38 \\
\cmark & \xmark & \cmark & 7.32  & 6.18  & 0.23 \\
\cmark & \cmark & \cmark & \textbf{6.79}  &\textbf{ 5.62} & \textbf{0.16} \\
\bottomrule
\end{tabular}
\caption{Ablation of \emph{Gate}, \emph{Local Align}, and \emph{Submap}. \cmark/\xmark  indicate module enabled/disabled. Using all three modules yields the best performance.}
\label{tab:ablation}
\end{table}

\subsection{Ablation Study}

In this section, we conduct various experiments to verify the effectiveness of our method. Table~\ref{tab:ablation} illustrates a quantitative evaluation with different settings.
We conduct an ablation study by respectively removing the gated update model, the 
submap scene representation, and the local alignment to prove the effectiveness of our modules. In comparison, both the submap representation and the state update model significantly contribute to the accuracy of scene reconstruction. Moreover, the hierarchical alignment plays a crucial role in enhancing tracking performance and maintaining overall consistency. As shown in Figure~\ref{fig:ablation}, we present an ablation study of the gated recurrent model without local alignment. Without the gated model, it fails to effectively regulate its latent state, resulting in the accumulation of inconsistent geometry and a loss of structural coherence. In contrast, our gated update method enables the selective integration of new observations, preserving spatial consistency across frames.

\section{Conclusion}

In this paper, we present a real-time dense end-to-end SLAM framework that achieves accurate scene reconstruction and camera tracking. We design a persistent latent state that serves as a spatial memory, continuously interacting with the current frame to update both its point cloud and pose, while simultaneously refining the memory itself. To prevent noise accumulation in the spatial memory and to improve the decoding of the current frame’s geometry, we introduce transformer-based update and reset gates that selectively control information flow. Furthermore, we propose a multi-submap representation strategy combined with a hierarchical alignment mechanism, which aligns submaps in a globally consistent manner while locally optimizing point clouds and poses within each submap. Experimental results on extensive datasets verify the effectiveness of our method.

%%%%%%%%%%%%%%%%%%%%%%%%%%%%%%%%%%%%%%%%%%%%%%%%%%%%%%%%%%%%%%%%%%%%%%%%%%%%%%%%

\bibliographystyle{ieeetran}
\bibliography{main}

\begin{thebibliography}{10}
\providecommand{\url}[1]{#1}
\csname url@rmstyle\endcsname
\providecommand{\newblock}{\relax}
\providecommand{\bibinfo}[2]{#2}
\providecommand\BIBentrySTDinterwordspacing{\spaceskip=0pt\relax}
\providecommand\BIBentryALTinterwordstretchfactor{4}
\providecommand\BIBentryALTinterwordspacing{\spaceskip=\fontdimen2\font plus
\BIBentryALTinterwordstretchfactor\fontdimen3\font minus \fontdimen4\font\relax}
\providecommand\BIBforeignlanguage[2]{{%
\expandafter\ifx\csname l@#1\endcsname\relax
\typeout{** WARNING: IEEEtran.bst: No hyphenation pattern has been}%
\typeout{** loaded for the language `#1'. Using the pattern for}%
\typeout{** the default language instead.}%
\else
\language=\csname l@#1\endcsname
\fi
#2}}

\bibitem{orbslam2}
R.~Mur-Artal and J.~D. Tardós, ``Orb-slam2: An open-source slam system for monocular, stereo, and rgb-d cameras,'' \emph{IEEE Transactions on Robotics}, vol.~33, no.~5, pp. 1255--1262, 2017.

\bibitem{deng}
T.~Deng, H.~Xie, J.~Wang, and W.~Chen, ``Long-term visual simultaneous localization and mapping: Using a bayesian persistence filter-based global map prediction,'' \emph{IEEE Robotics \& Automation Magazine}, vol.~30, no.~1, pp. 36--49, 2023.

\bibitem{xie}
H.~Xie, T.~Deng, J.~Wang, and W.~Chen, ``Robust incremental long-term visual topological localization in changing environments,'' \emph{IEEE Transactions on Instrumentation and Measurement}, vol.~72, pp. 1--14, 2022.

\bibitem{dtam}
R.~A. Newcombe, S.~J. Lovegrove, and A.~J. Davison, ``Dtam: Dense tracking and mapping in real-time,'' in \emph{2011 international conference on computer vision}.\hskip 1em plus 0.5em minus 0.4em\relax IEEE, 2011, pp. 2320--2327.

\bibitem{Kintinuous}
T.~Whelan, H.~Johannsson, M.~Kaess, J.~J. Leonard, and J.~McDonald, ``Kintinuous: Spatially extended kinectfusion,'' in \emph{RSS Workshop on RGB-D: Advanced Reasoning with Depth Cameras}, 2012.

\bibitem{NeRF}
B.~Mildenhall, P.~P. Srinivasan, M.~Tancik, J.~T. Barron, R.~Ramamoorthi, and R.~Ng, ``Nerf: Representing scenes as neural radiance fields for view synthesis,'' in \emph{ECCV}, 2020.

\bibitem{3dgs}
B.~Kerbl, G.~Kopanas, T.~Leimk{\"u}hler, and G.~Drettakis, ``3d gaussian splatting for real-time radiance field rendering.'' \emph{ACM Trans. Graph.}, vol.~42, no.~4, pp. 139--1, 2023.

\bibitem{niceslam}
Z.~Zhu, S.~Peng, V.~Larsson, W.~Xu, H.~Bao, Z.~Cui, M.~R. Oswald, and M.~Pollefeys, ``Nice-slam: Neural implicit scalable encoding for slam,'' in \emph{CVPR}, June 2022, pp. 12\,786--12\,796.

\bibitem{plgslam}
T.~Deng, G.~Shen, T.~Qin, J.~Wang, W.~Zhao, J.~Wang, D.~Wang, and W.~Chen, ``Plgslam: Progressive neural scene represenation with local to global bundle adjustment,'' in \emph{Proceedings of the IEEE/CVF Conference on Computer Vision and Pattern Recognition}, 2024, pp. 19\,657--19\,666.

\bibitem{neslam}
T.~Deng, Y.~Wang, H.~Xie, H.~Wang, R.~Guo, J.~Wang, D.~Wang, and W.~Chen, ``Neslam: Neural implicit mapping and self-supervised feature tracking with depth completion and denoising,'' \emph{IEEE Transactions on Automation Science and Engineering}, pp. 1--1, 2025.

\bibitem{snislam}
S.~Zhu, G.~Wang, H.~Blum, J.~Liu, L.~Song, M.~Pollefeys, and H.~Wang, ``Sni-slam: Semantic neural implicit slam,'' in \emph{Proceedings of the IEEE/CVF Conference on Computer Vision and Pattern Recognition}, 2024, pp. 21\,167--21\,177.

\bibitem{deng2025mcnslammultiagentcollaborativeneural}
T.~Deng, G.~Shen, X.~Chen, S.~Yuan, H.~Shen, G.~Peng, Z.~Wu, J.~Wang, L.~Xie, D.~Wang, H.~Wang, and W.~Chen, ``Mcn-slam: Multi-agent collaborative neural slam with hybrid implicit neural scene representation,'' \emph{arXiv preprint arXiv:2506.18678}, 2025.

\bibitem{gsslam}
C.~Yan, D.~Qu, D.~Wang, D.~Xu, Z.~Wang, B.~Zhao, and X.~Li, ``Gs-slam: Dense visual slam with 3d gaussian splatting,'' \emph{arXiv preprint arXiv:2311.11700}, 2023.

\bibitem{splatam}
N.~Keetha, J.~Karhade, K.~M. Jatavallabhula, G.~Yang, S.~Scherer, D.~Ramanan, and J.~Luiten, ``Splatam: Splat track \& map 3d gaussians for dense rgb-d slam,'' in \emph{Proceedings of the IEEE/CVF Conference on Computer Vision and Pattern Recognition}, 2024, pp. 21\,357--21\,366.

\bibitem{compact}
T.~Deng, Y.~Chen, L.~Zhang, J.~Yang, S.~Yuan, D.~Wang, and W.~Chen, ``Compact 3d gaussian splatting for dense visual slam,'' \emph{arXiv preprint arXiv:2403.11247}, 2024.

\bibitem{sgsslam}
M.~Li, S.~Liu, H.~Zhou, G.~Zhu, N.~Cheng, T.~Deng, and H.~Wang, ``Sgs-slam: Semantic gaussian splatting for neural dense slam,'' in \emph{European Conference on Computer Vision}.\hskip 1em plus 0.5em minus 0.4em\relax Springer, 2024, pp. 163--179.

\bibitem{mgslam}
S.~Liu, H.~Zhou, L.~Li, Y.~Liu, T.~Deng, Y.~Zhou, and M.~Li, ``Structure gaussian slam with manhattan world hypothesis,'' \emph{arXiv preprint arXiv:2405.20031}, 2024.

\bibitem{dust3r}
S.~Wang, V.~Leroy, Y.~Cabon, B.~Chidlovskii, and J.~Revaud, ``Dust3r: Geometric 3d vision made easy,'' in \emph{Proceedings of the IEEE/CVF Conference on Computer Vision and Pattern Recognition}, 2024, pp. 20\,697--20\,709.

\bibitem{cut3r}
Q.~Wang, Y.~Zhang, A.~Holynski, A.~A. Efros, and A.~Kanazawa, ``Continuous 3d perception model with persistent state,'' \emph{arXiv preprint arXiv:2501.12387}, 2025.

\bibitem{spann3r}
H.~Wang and L.~Agapito, ``3d reconstruction with spatial memory,'' \emph{arXiv preprint arXiv:2408.16061}, 2024.

\bibitem{slam3r}
Y.~Liu, S.~Dong, S.~Wang, Y.~Yin, Y.~Yang, Q.~Fan, and B.~Chen, ``Slam3r: Real-time dense scene reconstruction from monocular rgb videos,'' \emph{arXiv preprint arXiv:2412.09401}, 2024.

\bibitem{mast3rslam}
R.~Murai, E.~Dexheimer, and A.~J. Davison, ``Mast3r-slam: Real-time dense slam with 3d reconstruction priors,'' \emph{arXiv preprint arXiv:2412.12392}, 2024.

\bibitem{lsd-slam}
J.~Engel, T.~Sch{\"o}ps, and D.~Cremers, ``Lsd-slam: Large-scale direct monocular slam,'' in \emph{European conference on computer vision}.\hskip 1em plus 0.5em minus 0.4em\relax Springer, 2014, pp. 834--849.

\bibitem{cnn-slam}
K.~Tateno, F.~Tombari, I.~Laina, and N.~Navab, ``Cnn-slam: Real-time dense monocular slam with learned depth prediction,'' in \emph{Proceedings of the IEEE conference on computer vision and pattern recognition}, 2017, pp. 6243--6252.

\bibitem{codeslam}
M.~Bloesch, J.~Czarnowski, R.~Clark, S.~Leutenegger, and A.~J. Davison, ``Codeslam — learning a compact, optimisable representation for dense visual slam,'' in \emph{Proceedings of the IEEE Conference on Computer Vision and Pattern Recognition (CVPR)}, June 2018.

\bibitem{droidslam}
Z.~Teed and J.~Deng, ``Droid-slam: Deep visual slam for monocular, stereo, and rgb-d cameras,'' \emph{Advances in neural information processing systems}, vol.~34, pp. 16\,558--16\,569, 2021.

\bibitem{tandem}
L.~Koestler, N.~Yang, N.~Zeller, and D.~Cremers, ``Tandem: Tracking and dense mapping in real-time using deep multi-view stereo,'' in \emph{Conference on Robot Learning}.\hskip 1em plus 0.5em minus 0.4em\relax PMLR, 2022, pp. 34--45.

\bibitem{pixelperfectsfm}
P.~Lindenberger, P.-E. Sarlin, V.~Larsson, and M.~Pollefeys, ``Pixel-perfect structure-from-motion with featuremetric refinement,'' in \emph{Proceedings of the IEEE/CVF international conference on computer vision}, 2021, pp. 5987--5997.

\bibitem{Buildingromeinaday}
A.~Sameer, ``Building rome in a day,'' \emph{Proc. ICCV, 2009}, 2009.

\bibitem{sfmrevisited}
J.~L. Schonberger and J.-M. Frahm, ``Structure-from-motion revisited,'' in \emph{Proceedings of the IEEE conference on computer vision and pattern recognition}, 2016, pp. 4104--4113.

\bibitem{optimizingsfm}
C.~Sweeney, T.~Sattler, T.~Hollerer, M.~Turk, and M.~Pollefeys, ``Optimizing the viewing graph for structure-from-motion,'' in \emph{Proceedings of the IEEE international conference on computer vision}, 2015, pp. 801--809.

\bibitem{globalsfmrevisit}
L.~Pan, D.~Bar{\'a}th, M.~Pollefeys, and J.~L. Sch{\"o}nberger, ``Global structure-from-motion revisited,'' in \emph{European Conference on Computer Vision}.\hskip 1em plus 0.5em minus 0.4em\relax Springer, 2024, pp. 58--77.

\bibitem{Phototourism}
N.~Snavely, S.~M. Seitz, and R.~Szeliski, ``Photo tourism: exploring photo collections in 3d,'' in \emph{ACM siggraph 2006 papers}, 2006, pp. 835--846.

\bibitem{mvsnet}
Y.~Yao, Z.~Luo, S.~Li, T.~Fang, and L.~Quan, ``Mvsnet: Depth inference for unstructured multi-view stereo,'' in \emph{Proceedings of the European conference on computer vision (ECCV)}, 2018, pp. 767--783.

\bibitem{mvstutorial}
Y.~Furukawa, C.~Hern{\'a}ndez, \emph{et~al.}, ``Multi-view stereo: A tutorial,'' \emph{Foundations and trends{\textregistered} in Computer Graphics and Vision}, vol.~9, no. 1-2, pp. 1--148, 2015.

\bibitem{MVSeval}
S.~M. Seitz, B.~Curless, J.~Diebel, D.~Scharstein, and R.~Szeliski, ``A comparison and evaluation of multi-view stereo reconstruction algorithms,'' in \emph{2006 IEEE computer society conference on computer vision and pattern recognition (CVPR'06)}, vol.~1.\hskip 1em plus 0.5em minus 0.4em\relax IEEE, 2006, pp. 519--528.

\bibitem{accuratemvs}
Y.~Furukawa and J.~Ponce, ``Accurate, dense, and robust multiview stereopsis,'' \emph{IEEE transactions on pattern analysis and machine intelligence}, vol.~32, no.~8, pp. 1362--1376, 2009.

\bibitem{pixelwisemvs}
J.~L. Sch{\"o}nberger, E.~Zheng, J.-M. Frahm, and M.~Pollefeys, ``Pixelwise view selection for unstructured multi-view stereo,'' in \emph{Computer Vision--ECCV 2016: 14th European Conference, Amsterdam, The Netherlands, October 11-14, 2016, Proceedings, Part III 14}.\hskip 1em plus 0.5em minus 0.4em\relax Springer, 2016, pp. 501--518.

\bibitem{superpoint}
D.~DeTone, T.~Malisiewicz, and A.~Rabinovich, ``Superpoint: Self-supervised interest point detection and description,'' in \emph{Proceedings of the IEEE Conference on Computer Vision and Pattern Recognition (CVPR) Workshops}, June 2018.

\bibitem{ba-net}
C.~Tang and P.~Tan, ``Ba-net: Dense bundle adjustment network,'' \emph{ICLR}, 2018.

\bibitem{unidepth}
L.~Piccinelli, Y.-H. Yang, C.~Sakaridis, M.~Segu, S.~Li, L.~Van~Gool, and F.~Yu, ``Unidepth: Universal monocular metric depth estimation,'' in \emph{Proceedings of the IEEE/CVF Conference on Computer Vision and Pattern Recognition}, 2024, pp. 10\,106--10\,116.

\bibitem{zoedepth}
S.~F. Bhat, R.~Birkl, D.~Wofk, P.~Wonka, and M.~M{\"u}ller, ``Zoedepth: Zero-shot transfer by combining relative and metric depth,'' \emph{arXiv preprint arXiv:2302.12288}, 2023.

\bibitem{megadepth}
Z.~Li and N.~Snavely, ``Megadepth: Learning single-view depth prediction from internet photos,'' in \emph{Proceedings of the IEEE conference on computer vision and pattern recognition}, 2018, pp. 2041--2050.

\bibitem{diggingmonodepth}
C.~Godard, O.~Mac~Aodha, M.~Firman, and G.~J. Brostow, ``Digging into self-supervised monocular depth estimation,'' in \emph{Proceedings of the IEEE/CVF international conference on computer vision}, 2019, pp. 3828--3838.

\bibitem{mast3r}
V.~Leroy, Y.~Cabon, and J.~Revaud, ``Grounding image matching in 3d with mast3r,'' in \emph{European Conference on Computer Vision}.\hskip 1em plus 0.5em minus 0.4em\relax Springer, 2024, pp. 71--91.

\bibitem{vit}
A.~Dosovitskiy, L.~Beyer, A.~Kolesnikov, D.~Weissenborn, X.~Zhai, T.~Unterthiner, M.~Dehghani, M.~Minderer, G.~Heigold, S.~Gelly, \emph{et~al.}, ``An image is worth 16x16 words: Transformers for image recognition at scale,'' \emph{arXiv preprint arXiv:2010.11929}, 2020.

\bibitem{GRU}
J.~Chung, C.~Gulcehre, K.~Cho, and Y.~Bengio, ``Empirical evaluation of gated recurrent neural networks on sequence modeling,'' \emph{arXiv preprint arXiv:1412.3555}, 2014.

\bibitem{dpt}
R.~Ranftl, A.~Bochkovskiy, and V.~Koltun, ``Vision transformers for dense prediction,'' in \emph{Proceedings of the IEEE/CVF international conference on computer vision}, 2021, pp. 12\,179--12\,188.

\bibitem{co3dv2}
J.~Reizenstein, R.~Shapovalov, P.~Henzler, L.~Sbordone, P.~Labatut, and D.~Novotny, ``Common objects in 3d: Large-scale learning and evaluation of real-life 3d category reconstruction,'' in \emph{Proceedings of the IEEE/CVF international conference on computer vision}, 2021, pp. 10\,901--10\,911.

\bibitem{arkitscenes}
A.~Dehghan, G.~Baruch, Z.~Chen, Y.~Feigin, P.~Fu, T.~Gebauer, D.~Kurz, T.~Dimry, B.~Joffe, A.~Schwartz, \emph{et~al.}, ``Arkitscenes: A diverse real-world dataset for 3d indoor scene understanding using mobile rgb-d data.'' \emph{NeurIPS Datasets and Benchmarks}, vol.~2, no.~6, p.~16, 2021.

\bibitem{scannet}
A.~Dai, A.~X. Chang, M.~Savva, M.~Halber, T.~Funkhouser, and M.~Niessner, ``Scannet: Richly-annotated 3d reconstructions of indoor scenes,'' in \emph{Proceedings of the IEEE Conference on Computer Vision and Pattern Recognition (CVPR)}, July 2017.

\bibitem{wildrgbd}
H.~Xia, Y.~Fu, S.~Liu, and X.~Wang, ``Rgbd objects in the wild: scaling real-world 3d object learning from rgb-d videos,'' in \emph{Proceedings of the IEEE/CVF Conference on Computer Vision and Pattern Recognition}, 2024, pp. 22\,378--22\,389.

\bibitem{blendedmvs}
Y.~Yao, Z.~Luo, S.~Li, J.~Zhang, Y.~Ren, L.~Zhou, T.~Fang, and L.~Quan, ``Blendedmvs: A large-scale dataset for generalized multi-view stereo networks,'' in \emph{Proceedings of the IEEE/CVF conference on computer vision and pattern recognition}, 2020, pp. 1790--1799.

\bibitem{matterport3d}
A.~Chang, A.~Dai, T.~Funkhouser, M.~Halber, M.~Niessner, M.~Savva, S.~Song, A.~Zeng, and Y.~Zhang, ``Matterport3d: Learning from rgb-d data in indoor environments,'' \emph{arXiv preprint arXiv:1709.06158}, 2017.

\bibitem{neural_rgbd}
D.~Azinovi\'c, R.~Martin-Brualla, D.~B. Goldman, M.~Nie{\ss}ner, and J.~Thies, ``Neural rgb-d surface reconstruction,'' in \emph{Proceedings of the IEEE/CVF Conference on Computer Vision and Pattern Recognition (CVPR)}, June 2022, pp. 6290--6301.

\bibitem{7scenes}
J.~Shotton, B.~Glocker, C.~Zach, S.~Izadi, A.~Criminisi, and A.~Fitzgibbon, ``Scene coordinate regression forests for camera relocalization in rgb-d images,'' in \emph{Proceedings of the IEEE conference on computer vision and pattern recognition}, 2013, pp. 2930--2937.

\bibitem{replica}
J.~Straub, T.~Whelan, L.~Ma, Y.~Chen, E.~Wijmans, S.~Green, J.~J. Engel, R.~Mur-Artal, C.~Ren, S.~Verma, \emph{et~al.}, ``The replica dataset: A digital replica of indoor spaces,'' \emph{arXiv preprint arXiv:1906.05797}, 2019.

\bibitem{deng2025mne}
T.~Deng, G.~Shen, C.~Xun, S.~Yuan, T.~Jin, H.~Shen, Y.~Wang, J.~Wang, H.~Wang, D.~Wang, \emph{et~al.}, ``Mne-slam: Multi-agent neural slam for mobile robots,'' in \emph{Proceedings of the Computer Vision and Pattern Recognition Conference}, 2025, pp. 1485--1494.

\bibitem{nicerslam}
Z.~Zhu, S.~Peng, V.~Larsson, Z.~Cui, M.~R. Oswald, A.~Geiger, and M.~Pollefeys, ``Nicer-slam: Neural implicit scene encoding for rgb slam,'' in \emph{2024 International Conference on 3D Vision (3DV)}.\hskip 1em plus 0.5em minus 0.4em\relax IEEE, 2024, pp. 42--52.

\end{thebibliography}

\end{document}